\documentclass[10pt,twocolumn,letterpaper]{article}

\usepackage{cvpr}              

\usepackage{amsmath,amsfonts,bm}

\def\eqref#1{equation~\ref{#1}}

\def\1{\bm{1}}

\def\vs{{\bm{s}}}

\DeclareMathAlphabet{\mathsfit}{\encodingdefault}{\sfdefault}{m}{sl}
\SetMathAlphabet{\mathsfit}{bold}{\encodingdefault}{\sfdefault}{bx}{n}

\usepackage{graphicx}
\usepackage{amsmath}
\usepackage{amssymb}
\usepackage{mathrsfs}
\usepackage{mathtools}
\usepackage{booktabs}
\usepackage{rotating}
\usepackage{multirow}
\usepackage{lipsum}
\usepackage{xcolor}
\usepackage{enumitem}
\usepackage{inconsolata}
\usepackage{pifont}
\usepackage{soul}
\usepackage{siunitx}
\usepackage{csquotes}
\usepackage[accsupp]{axessibility}  

\usepackage{microtype}      
\usepackage{threeparttable}

\usepackage{subcaption}

\usepackage[table]{xcolor}

\definecolor{scorelow}{HTML}{E0F2F1}  
\definecolor{scoremed}{HTML}{B2DFDB}
\definecolor{scorehigh}{HTML}{4DB6AC}
\definecolor{scorevhigh}{HTML}{00897B} 
\definecolor{humanperf}{gray}{0.9}     

\newcommand{\setcellbg}[1]{%
  \ifdim #1pt > 55pt \cellcolor{scorevhigh!90} \else
  \ifdim #1pt > 45pt \cellcolor{scorehigh!80} \else
  \ifdim #1pt > 30pt \cellcolor{scoremed!70} \else
  \ifdim #1pt > 15pt \cellcolor{scorelow!60} \else
  \cellcolor{white} 
  \fi\fi\fi\fi
}

\newcolumntype{C}{>{\centering\arraybackslash}X}

\newcommand{\todohs}[1]{{\color{red} HS: #1}}
\renewcommand{\todohs}[1]{}

\usepackage{amsfonts}
\usepackage{nicefrac}       
\usepackage{tabularx}
\usepackage{array}          
\usepackage{longtable}
\usepackage{wrapfig}
\usepackage{tablefootnote}
\usepackage{caption}
\usepackage[dvipsnames]{xcolor}
\usepackage{xspace}
\usepackage{comment}
\usepackage{adjustbox}
\usepackage{algorithm}
\usepackage{algpseudocode}
\usepackage{fontawesome}
\usepackage{marvosym}
\usepackage{colortbl}
\usepackage[framemethod=TikZ]{mdframed}

\usepackage[normalem]{ulem}

\definecolor{cvprblue}{rgb}{0.21,0.49,0.74}
\usepackage[pagebackref,breaklinks,colorlinks,allcolors=cvprblue]{hyperref}

\definecolor{velopink}{RGB}{190, 60, 108}

\definecolor{veloblue}{RGB}{83, 149, 218}
\definecolor{veloorange}{RGB}{236, 133, 45}

\definecolor{actioncolor}{RGB}{79, 173, 91}
\definecolor{agentcolor}{RGB}{104, 51, 154}
\definecolor{meucolor}{RGB}{222, 48, 135}

\makeatletter
\def\thanks#1{\protected@xdef\@thanks{\@thanks
        \protect\footnotetext{#1}}}
\makeatother

\definecolor{strictvlecolor}{RGB}{246, 246, 210}
\definecolor{scatcorrect}{RGB}{166, 227, 136}
\definecolor{scatstrict}{RGB}{46, 77, 27}
\definecolor{scatincorrect}{RGB}{203, 59, 39}

\definecolor{qablue}{RGB}{190, 214, 234}

\definecolor{forestgreen}{rgb}{0.13, 0.55, 0.13}
\definecolor{fireenginered}{rgb}{0.81, 0.09, 0.13}
\definecolor{graydash}{RGB}{150,150,150}
\definecolor{darkred}{rgb}{0.6, 0.1, 0.1}
\definecolor{lightgreen}{HTML}{E6F4EA}
\definecolor{myyellow}{HTML}{FBBC05}

\newcommand{\cmark}{{\textcolor{forestgreen}{\ding{51}}}}%

\usepackage{tikz}

\definecolor{ultralightgray}{RGB}{245,245,245}  

\renewcommand{\emph}[1]{\textit{#1}}
\renewcommand{\paragraph}[1]{\vspace{0.8mm}\noindent\textbf{#1}}

\newcommand{\bbf}{\mathbf{f}}

\newcommand{\bh}{\mathbf{h}}

\newcommand{\CLS}{\mathsf{CLS}}

\newcommand{\MN}{\mathcal{N}}

\newcommand{\ML}{\mathcal{L}}

\algrenewcommand\algorithmicrequire{\textbf{Input:}}
\algrenewcommand\algorithmicensure{\textbf{Output:}}

\newmdenv[backgroundcolor=cyan!10, linecolor=black, linewidth=0.5pt, roundcorner=4pt, innerleftmargin=2pt, innerrightmargin=2pt, innertopmargin=5pt, innerbottommargin=5pt]{exampleprompt}

\newcommand{\modelname}{SRL-CLIP}

\newcommand{\cgray}[1]{{\color{gray} #1}}
\definecolor{deepred}{rgb}{0.9, 0.17, 0.31} 
\newcommand{\cdeepred}[1]{{\color{deepred} #1}}
\definecolor{myblue}{rgb}{0.0, 0.28, 0.67} 
\newcommand{\cblue}[1]{{\color{myblue} #1}}

\definecolor{cyan}{rgb}{0.0, 1.0, 1.0}

\renewcommand{\paragraph}[1]{\vspace{1mm}\noindent\textbf{#1}}
\newcommand{\bt}{\mathbf{t}}
\newcommand{\bv}{\mathbf{v}}

\newcommand{\be}{\mathbf{e}}

\newcommand{\bEtype}{\mathbf{e}^{\text{typ}}}
\newcommand{\bEepos}{\mathbf{e}^{\text{e-pos}}}
\newcommand{\bEfpos}{\mathbf{e}^{\text{f-pos}}}

\newcommand{\mean}{\text{mean}}

\definecolor{deepred}{rgb}{0.9, 0.17, 0.31} 

\definecolor{myblue}{rgb}{0.0, 0.28, 0.67} 

\definecolor{cyan}{rgb}{0.0, 1.0, 1.0}

\RequirePackage{xspace}
\makeatletter
\DeclareRobustCommand\onedot{\futurelet\@let@token\@onedot}
\def\@onedot{\ifx\@let@token.\else.\null\fi\xspace}

\def\eg{\emph{e.g}\onedot}
\def\Eg{\emph{E.g}\onedot}
\def\ie{\emph{i.e}\onedot}

\def\etc{\emph{etc}\onedot}
\def\vs{\emph{vs}\onedot}

\makeatother

\AtEndPreamble{
    \usepackage[capitalize]{cleveref}
    \crefname{section}{Sec.}{Secs.}
    \Crefname{section}{Section}{Sections}
    \crefname{table}{Tab.}{Tabs.}
    \Crefname{table}{Table}{Tables}
}

\usepackage[accsupp]{axessibility}  

\definecolor{cvprblue}{rgb}{0.21,0.49,0.74}
\usepackage[pagebackref,breaklinks,colorlinks,allcolors=cvprblue]{hyperref}
\usepackage{multirow}

\usepackage{booktabs}   
\usepackage{graphicx}   
\usepackage{makecell}   
\usepackage{pifont}     
\usepackage{epigraph}
\def\paperID{***} 
\def\confName{CVPR}
\def\confYear{2026}

\usepackage[colorlinks]{hyperref}
\usepackage{url}
\usepackage{caption}
\usepackage{color, colortbl}
\definecolor{LightCyan}{rgb}{0.88,1,1}

\title{SRL-CLIP: Efficient CLIP Video Adaptation via Structured Semantic Role Labels}

\author{
Darshan Singh$^1$,
Zeeshan Khan$^2$,
Makarand Tapaswi$^1$
\\
\small $^1$CVIT, IIIT Hyderabad \qquad $^2$Inria, École normale supérieure, CNRS, PSL Research University
\\
}

\begin{document}

\maketitle

\begin{abstract}
\vspace{-2mm}

Adapting CLIP for videos has gained popularity due to its semantic and rich representation.
While CLIP is a good starting point, it typically undergoes post-pretraining (contrastive finetuning) on large video narration or caption datasets (\eg~HowTo100M, WebVid2.5M).
However, such narrations or captions often lack comprehensive information needed to represent a video holistically.
As the learning signal from text is sparse, the visual learning is inefficient and adaptation requires millions of samples to post-pretrain.

In this work, we ask: is it possible to \textit{efficiently} adapt CLIP for general and holistic video understanding?
We use videos labeled with structured and dense \textit{Semantic Role Labels} (SRLs) that capture actions, people or objects, their attributes, adverbs (manner), and location in a structured format representing the entire video in a holistic way.
We generate rule-based captions from SRLs and demonstrate that simple contrastive finetuning on a \textbf{mere 23k} video-caption pairs is adequate to learn powerful, transferable representations applicable across a diverse range of video understanding tasks that require varying levels of perceptual granularity.
Our adapted CLIP model, \modelname, exhibits comparable or superior performance on zero-shot text-to-video retrieval compared to state-of-the-art models that possess $4{-}8\times$ more parameters and are post-pretrained on up to $6000\times$ more data.
\modelname{} surpasses CLIP on multiple video benchmarks, underscoring the efficient learning and improved representations.

\end{abstract}

\section{Introduction}
\label{sec:intro}

Large-scale vision language pretraining has proved effective in learning generalized and transferable visual representations, with powerful zero-shot capabilities~\citep{clip,blip,blipv2,albef}.
Among them, Contrastive Language-Image Pretraining (CLIP)~\citep{clip} is a popular approach for learning rich semantic representations, thanks to the image-text alignment.
As the representations generalize well, CLIP sees wide adoption in several downstream tasks ranging from simple classification to being used in Vision-Language Models (VLMs)~\citep{flamingo, pixelllm, cogvlm, instructblip}.
This has spurred a lot of interest in adapting CLIP for video understanding~\citep{clip4clip, verbsinaction, vificlip, bt_adapter}.
While these methods achieve promising results, they often come at the expense of large scale post-pretraining on hundreds of millions of videos.

We argue that as CLIP is pretrained on 400+ million image-language pairs, it already has the world knowledge required for general visual perception, and post-pretraining on hundreds of millions of videos can be wasteful.
We hypothesize that the need for such large-scale post-pretraining is due to the contradiction between rich videos and sparse descriptions.
Videos are a highly complex modality with multidimensional information about people, objects, states, actions, relations, that change over time.
But, the textual description representing the video is generally very sparse.
Narrations~\citep{howto100m} or captions~\citep{webvid2.5m} often fail to capture details and therefore the capacity for visual processing is underutilized.
The adaptation process requires millions of samples as it only extracts sparse information from each.

\begin{figure}[t]
\centering
\includegraphics[width=\linewidth]{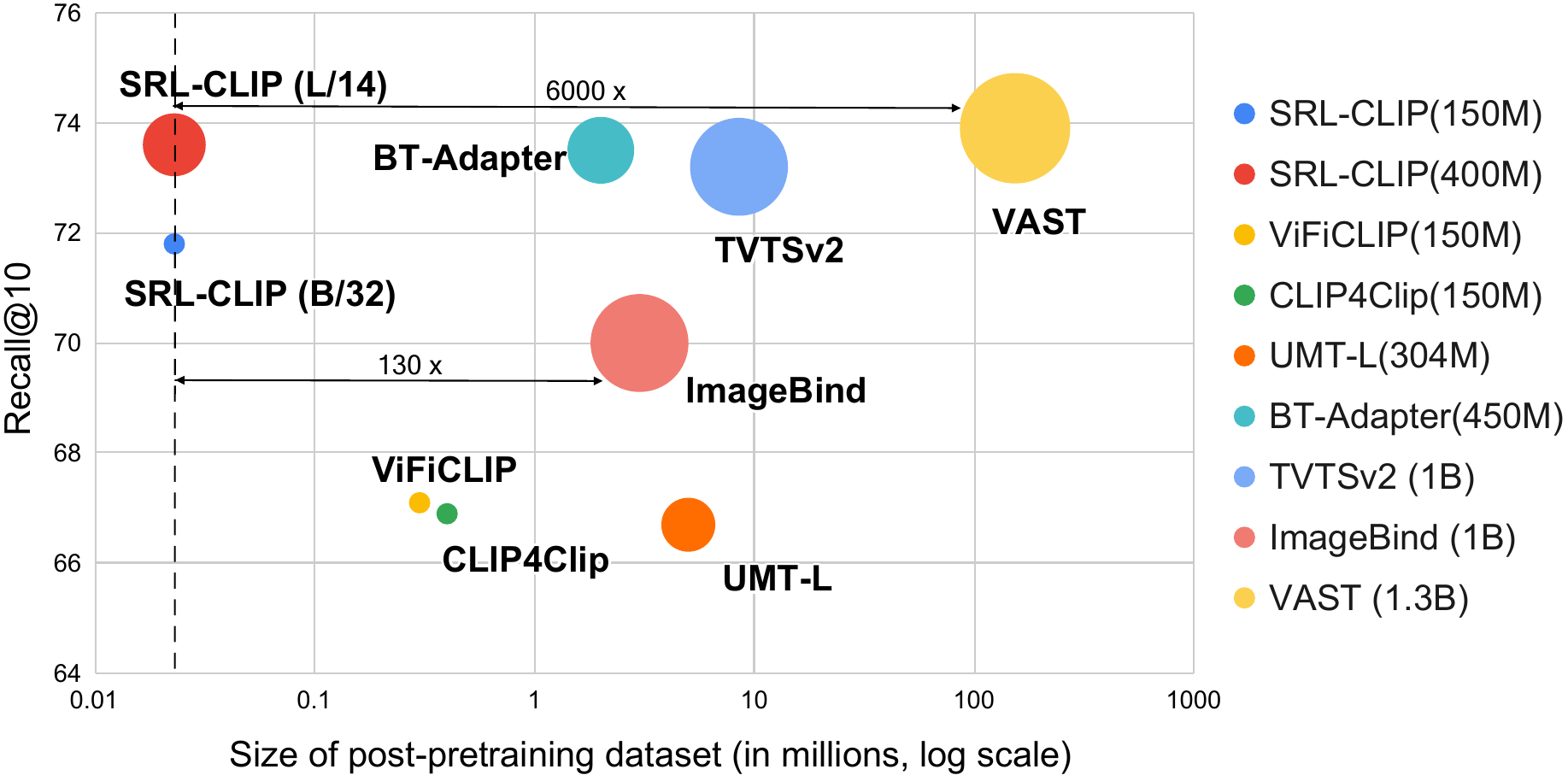}
\vspace{-5mm}
\caption{Zero-shot text-to-video retrieval performance on the MSRVTT dataset.
We compare \modelname{} (Ours) against various CLIP-based approaches that use orders of magnitude more post-pretraining data samples and/or have significantly larger models.}
\vspace{-5mm}
\label{fig:msrvtt_teaser}
\end{figure}

In this work, we revisit the large-scale post-pretaining paradigm for CLIP video adaptation and propose to use small-scale, but dense captions for efficient and holistic representation learning.
Specifically, we use the VidSitu dataset~\citep{vidsitu} that is annotated using Semantic Role Labels (SRLs) that capture the holistic situation.
In VidSitu, every 10 second video clip is divided into 2 second \emph{events}.
Each event contains structured annotations answering who (subject) is doing what (action), to/with whom (patient), where (scene), how (manner, adverbs), and why (purpose, goal).
Such details in a video description represent the visual concepts well and can provide a strong learning signal.

We propose to adapt CLIP through alignment between video-text SRLs, highlighting the power of rule-based high quality prompts generated from SRL.
Our approach has multiple advantages:
(i)~The structured nature of SRL-based prompts (who, what, where, \etc) facilitate learning holistic visual concepts consistently across all the videos.
(ii)~VidSitu videos have many shot changes, while the annotated captions may represent visual concepts across all the events.
This presents a challenging scenario with a strong signal to learn object permanence and temporal consistency, by meaningfully combining information across frames and events.
(iii)~Multiple events from the same videos show small differences creating \emph{natural hard negatives} for learning. In addition, due to the structured nature of SRL, we can add/swap/replace verbs, nouns, or roles in an SRL-based prompt, easily creating \emph{hard negative text prompts} that further boost the model's performance.

We show that \modelname{}, adapted on a small dataset with dense annotations, learns powerful general purpose representations, improving over base CLIP on multiple video-language tasks.
In particular, we achieve these improvements through \emph{efficient adaptation} --
a \texttt{ViT-B/32} with 150M params,
post-pretraining dataset of 23k video clips,
trained on a single 12GB RTX 2080 GPU,
for only 5 hours!

\paragraph{Contributions.}
In summary, we:
(i)~Favor efficient post-pretraining using small-scale but dense prompts, as compared to the current inefficient trend of using large-scale datasets.
(ii)~Propose to use SRL to create rule-based dense captions that capture holistic concepts in a video, through details such as actions, persons, objects, attributes, relations, manner, location, reasons, \etc.
(iii)~Show that post-pretraining CLIP on mere 23k videos with dense SRL captions is efficient and achieves on-par or better performance on zero-shot text-to-video retrieval (see \cref{fig:msrvtt_teaser}), compared to CLIP based video models that have $4{-}8\times$ more parameters and are trained on upto $6000\times$ more data.
(iv)~Outperform the original CLIP model on various tasks:
video situation recognition,
dense video captioning and localization, and
text-to-video retrieval,
demonstrating that representations learned with dense holistic captions generalize well across multiple tasks requiring different perceptual granularity.

We emphasize that our innovation lies in an efficient and effective recipe to adapt CLIP for video data rather than architectural or modeling modifications.

\subsection{Small Data Statement}
\label{subsec:small_data_statement}

Our work qualifies as small data research because the entire post-pretraining of \modelname{} is performed on only \textit{23k video clips} from VidSitu~\citep{vidsitu}. This is in stark contrast to existing CLIP video adaptation methods that rely on datasets that are orders of magnitude larger than ours as shown in~\cref{table:zeroshot_t2v_main}. Our training is also carried out on a single 12GB RTX 2080 GPU in just 5 hours for the \texttt{ViT-B/32} model, achieving competitive or superior results to models with up to $4-8\times$ more parameters and trained on orders of magnitude more data.

To tackle the small data challenge, we employ the following methods:
(i)~\emph{Dense structured annotations}: We leverage Semantic Role Labels (SRLs) that capture who, what, whom, where, how, and why for each event, providing a much richer per-sample learning signal than sparse narrations or captions. This compensates for the small dataset size by maximizing the information extracted from each video.
(ii)~\emph{Parameter-efficient adaptation}: LoRA modules~\citep{lora} are added to a frozen CLIP backbone, preventing catastrophic forgetting while enabling effective adaptation with limited data.
(iii)~\emph{Hard negatives from structure}: The compositional nature of SRL allows us to systematically create hard negatives by swapping verbs, nouns, or roles, boosting fine-grained understanding without requiring additional data.
(iv)~\emph{Video contextualizer}: A lightweight temporal module contextualizes frame-level features across events, enabling fine-grained temporal understanding from limited videos.

\section{Related Work}
\label{sec:relwork} 

Several attempts have been made to adapt CLIP for videos~\citep{videoclip,actionclip,xclip,clip4clip,clipvip,vificlip,disspatemp,revisitvideo,clip2video,UCoFiA}.
However, they are primarily focused on task-specific adaptation for action recognition~\citep{actionclip, vificlip} or text-to-video retrieval~\citep{videoclip, clipvip, clip2video, clip4clip, xclip, UCoFiA}.
Task specific adaptations may lead to the loss of generalized representations.
Different from above, we propose task-agnostic adaptation of CLIP, and focus on extending CLIP to videos using holistic video understanding datasets.
This allows our model to be applied on a diverse range of video understanding tasks that demand different levels of perceptual granularity.
We discuss related works in two dimensions:
(i)~datasets used for adapting CLIP; and
(ii)~approaches for adapting CLIP for videos.

\paragraph{1. Popular VL datasets for adapting CLIP.}
The unparalleled success of training Large Language Models (LLMs), \eg~GPT2~\citep{gpt2}, LLaMA~\citep{llama}, with massive datasets
has brought similar trends to the vision community.

\paragraph{Towards text-to-video retrieval.}
Large-scale web scraped datasets such as HowTo100M~\citep{howto100m} and HD-VILA~\citep{hdvila} align video clips with narrations.
More recently, WebVid-2.5M~\citep{webvid2.5m} was curated from stock footage with textual descriptions resulting in better captions, that align with the video.
These datasets are used by several methods for adapting CLIP or training a video-text contrastive alignment approach from scratch~\citep{clip4clip,videoclip,clipvip,supportset,multitxvidret}.

\paragraph{Towards action understanding.}
Datasets typically consist of 10-second videos annotated with a single action.
To adapt CLIP using supervised video-text contrastive learning, the widely used idea is to create CLIP-like prompts from the action labels~\citep{vificlip,actionclip,xclip} on datasets like Kinetics~\citep{kinetics} or Something-Something~\citep{ss}.

\paragraph{Towards holistic video understanding.}
Action labels~\citep{kinetics} or narrations~\citep{howto100m} fail to capture the complex, hierarchical, multifaceted aspects of videos.
To learn holistic and fine-grained representations, we propose to exploit the SRL annotations in VidSitu~\citep{vidsitu}.
We observe that \modelname{} consistently outperforms base CLIP on multiple video understanding tasks that require varying degrees of granularity, indicating that it has improved holistic reasoning abilities.
Moreover, a model trained on SRL captions outperforms methods that are post-pretrained on large-scale web video datasets (often 2-3 orders of magnitude larger)~\citep{howto100m,hdvila,webvid2.5m} on \textit{zero-shot} video retrieval, indicating improved vision-language alignment.

\paragraph{2. Approaches for adapting CLIP} for videos
can be classified into direct fine-tuning or through adapters.

\paragraph{Fine-tuning approaches}
typically follow frame-level feature extraction from CLIP followed by temporal aggregation.
The resulting video feature is aligned with the corresponding text prompt via contrastive loss.
Either partial, or all the parameters of CLIP are fine-tuned~\citep{actionclip,vificlip}.
While a majority of the methods~\citep{actionclip,xclip,clip4clip,clip2video} use a Transformer~\citep{transformer} for temporal aggregation of frame-level features, simple mean pooling has been shown to be effective for a narrow task of action recognition~\citep{vificlip}.
Other approaches
use a weighted-mean of frame embeddings based on query-scoring~\citep{cliphitchhiker};
compute frame-level attention based on text~\citep{xpool};
integrate temporal aggregation within the image encoder~\citep{clipvip}; or
suggest using a temporal model in parallel to the image encoder~\citep{dissapttemp}.

\paragraph{Adapters}, the alternative to finetuning, are lightweight modules
injected between layers of a pretrained model for efficient adaptation on a downstream task~\citep{nlpadapters,lora,vladapter,svladapterbmvc2022}.
The original parameters are usually frozen, and only the adapter is trained, allowing for efficient adaptation.
Prior works in CLIP adaptation have used
spatial adapters~\citep{clipadapter2023ijcv},
spatio-temporal adapters~\citep{stadapter2022neurips}, and 
cross-modal adapters~\citep{crossadapter}
for efficient adaptation to downstream tasks.
Recently there has been a surge of methods using low-rank adapters (LoRA)~\citep{lora} and advances~\citep{loraplus, liu2024dora} 
for their high efficiency enabled by low rank learnable matrices during training and no computational overhead during inference.
We follow this approach instead of fine-tuning, allowing us to efficiently post-pretrain CLIP's \texttt{ViT-B/32} image and/or text encoders on a single 12GB GPU.

\section{Adaptation: From CLIP to \modelname}
\label{sec:method}

We present our approach to post-pretrain the CLIP model on VidSitu, a densely annotated video dataset.
We start with background information related to CLIP (\cref{subsec:method:prelims}), followed by our adaptation strategy with specific emphasis on the architecture modifications for training and SRL prompts (\cref{subsec:method:adapt_vidsitu}).
We end this section with a discussion on how SRL prompts support creation of difficult negatives (\cref{subsec:method:negatives}) and some details about the adaptation process (\cref{subsec:method:details}).

\subsection{Preliminaries}
\label{subsec:method:prelims}
Consider a batch $B$ of paired image-text data: $\{(f_i, t_i)\}_{i=1}^B$,
where $f_i$ is the image and $t_i$ describes $f_i$.
The CLIP model~\citep{clip} consists of an image encoder $\bbf_i = \Phi_I(f_i)$ and a text encoder $\bt_i = \Phi_T(t_i)$ that are trained in a contrastive manner by applying the InfoNCE loss~\citep{infonce}:
\begin{equation}
\label{eq:nce}
L(\bbf_i, \bt_i) = - \log \frac{\exp(\bbf_i^T \bt_i)}{\sum_{j=1}^B \exp(\bbf_i^T \bt_j)} \, ,
\end{equation}
and the corresponding symmetric version $L(\bt_i, \bbf_i)$.
The loss for the entire batch is $\ML = \sum_{i=1}^B ( L(\bbf_i, \bt_i) + L(\bt_i, \bbf_i) )$.

Previous works have adapted CLIP using video-text pairs by mean pooling across multiple video frames~\citep{vificlip}.
Instead of an image $f_i$, consider a video-text pair $(V_i, t_i)$ where $V_i = \{f_{ij}\}_{j=1}^{L_i}$ has $L_i$ frames.
We can adapt CLIP by computing a video representation $\bv_i = \mean_j( \bbf_{ij} )$ and using the same loss $L(\bv_i, \bt_i)$.

\paragraph{Video contextualizer (VC).}
Instead of mean pooling, a VC may be used for learning video-text representations~\citep{xclip,clip4clip}.
To contextualize all frames, a simple Transformer encoder (Tx)~\citep{transformer} ingests frame representations as tokens.
A learnable $\CLS$ token, $\bh_\CLS$, is inserted before all frames.
The input to the VC is:
$[\bh_\CLS, \bbf_{i1}, \ldots, \bbf_{iL_i}]$.
Position encoding~\citep{transformer} is added to the video frame tokens to specify their temporal order.
Finally, the output at the $\CLS$ token is considered as the video representation, \ie, $\bv_i = \hat{\bh}_\CLS$.
Note, the VC can be trained jointly with adaptation of the backbone through the same loss $L(\bv_i, \bt_i)$.

Next, we show how VC can be used with VidSitu.

\begin{figure}[t!]
\centering
\includegraphics[width=\linewidth]{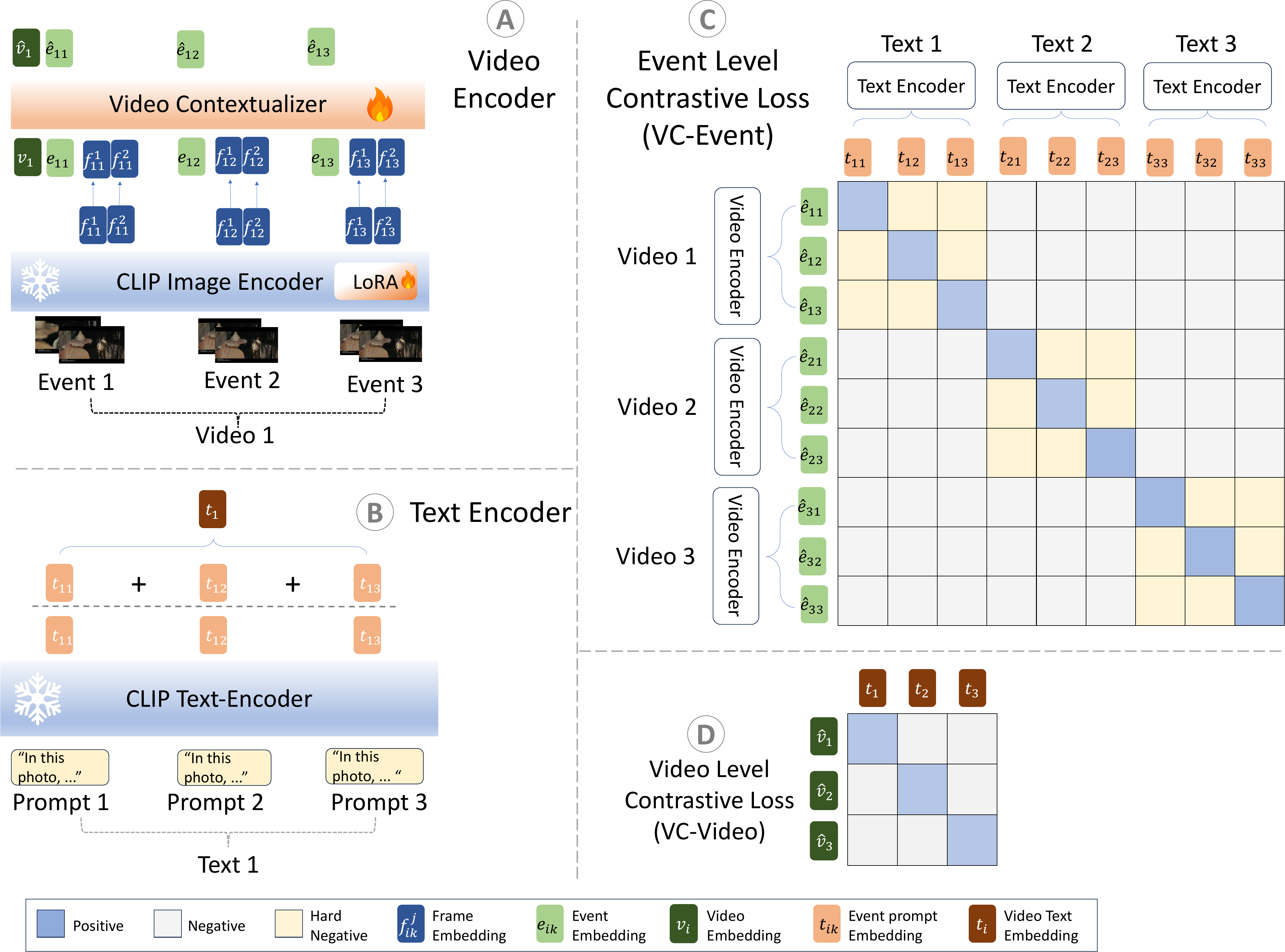}
\caption{Overview of our CLIP adaptation strategy.
\textbf{Top-left (A)} shows the visual encoder, consisting of the CLIP backbone and the video contextualizer (VC), applied to a single video with $P{=}3$ events for illustration.
\textbf{Bottom-left (B)} shows the frozen CLIP text encoder extracting event-level representations.
\textbf{Top-right (C)} shows the event-level contrastive loss, VC-Event, with natural hard negatives due to multiple events within a single video.
\textbf{Bottom-right (D)} shows the video-level contrastive loss, VC-Video.
}
\vspace{-2mm}
\label{fig:model}
\end{figure}

\subsection{Adaptation with SRLs}
\label{subsec:method:adapt_vidsitu}

The VidSitu dataset~\citep{vidsitu} features videos that are split into $P{=}5$ contiguous short \emph{events}, \ie~$V_i = [E_{ik}]_{k=1}^P$.
Each event contains a detailed annotation:
an action label and corresponding semantic role labels (SRL) with the role-noun pairs.
For example, in \cref{fig:main_paper_qualitative}, we show the action \emph{drive}, with roles \emph{driver}, \emph{vehicle}, \emph{manner} (of driving), and \emph{scene}, each described through a short caption (noun).

We use such fine-grained labels to create a prompt for each event, $t_{ik}$, leading to event and text pairs $(E_{ik}, t_{ik})$.
\cref{fig:model} illustrates the overall adaptation strategy.

\paragraph{Video contextualizer (VC) for encoding VidSitu events.}
We modify the VC to account for VidSitu's structured annotations.
During training, $B$ videos $\{V_i\}_{i=1}^B$ are fed to the model at once.
Each video is split into $P$ events.
From each event, we sub-sample $T$ frames, \ie, for each video, we have $L {=} P\cdot T$ frames.
The VC operates over a sequence of all frames, passed through the CLIP image encoder.
Let $f_{ik}^j$ be the $j^{\text{th}}$ frame for event $E_{ik}$ of video $V_i$.
$\bv_i$ represents the overall video $V_i$ and $\be_{ik}$ represents event $E_{ik}$.

Similar to $\CLS$ in BERT~\citep{bert}, we create two learnable token types that collect video information.
Note, these embeddings are shared across all videos.
To indicate the type of token, we augment visual/learnable encodings with type embeddings $\bEtype_v$ for video, $\bEtype_e$ for event, and $\bEtype_f$ for the frame.
Furthermore, we encode position with two embeddings, $\bEepos$ for event position and $\bEfpos$ for frame position within the event.
Overall, our input tokens are:
\begin{eqnarray}
\bv_i &=& \bv_i + \bEtype_v \, , \\
\be_{ik} &=& \be_{ik} + \bEtype_e + \bEepos_k \, , \\
\bbf_{ik}^j &=& \bbf_{ik}^j + \bEtype_f + \bEepos_k + \bEfpos_j \, ,
\end{eqnarray}
and passed to the VC, $\Phi_V$, after LayerNorm~\cite{layernorm}:
\begin{equation}
\Phi_V([\bv_i,
\be_{i1}, \bbf_{i1}^1, \ldots, \bbf_{i1}^T,
\ldots,
\be_{iP}, \bbf_{iP}^1, \ldots, \bbf_{iP}^T]) \, .
\end{equation}
We denote outputs after the VC as $\hat{\bv}_i, \hat{\be}_{ik}, \hat{\bbf}^j_{ik}$ for video, event, and frame tokens respectively.
\cref{fig:model} (top-left) illustrates this process.

\paragraph{Creating event-level prompts.}
Our prompt $t_{ik}$ is a simple template that enumerates over the action and semantic role labels for each event.
An example is shown below.
Template words in \cgray{gray},
label type in \emph{italics},
the label in \underline{line}:
\begin{displayquote}
\cgray{In this photo, the}
\emph{action} \cgray{is}
\underline{walk}
\cgray{where, the}
\emph{walker} \cgray{is}
\underline{man with short hair} \underline{wearing collared shirt},
\emph{direction} \cgray{is}
\underline{forward},
\emph{manner} \cgray{is}
\underline{slowly}, \cgray{and}
\emph{scene} \cgray{of the event is}
\underline{apartment}.
\end{displayquote}

We also consider generating natural language prompts using a language model (LLaMa~\citep{llama}).
Labels are \underline{underlined}:
\begin{displayquote}
\cgray{In this photo,}
\underline{a man with short hair wearing a}
\underline{collared shirt}
\cgray{is}
\underline{walking} \underline{slowly}
\cgray{in an}
\underline{apartment}.
\end{displayquote}
However, these show worse performance.
Telling the model that walking is the \emph{action}, or the role played by a person with collared shirt is \emph{walker}, and the apartment is a \emph{scene} allows for holistic and dense representation learning.

\paragraph{Losses.}
We train our model with losses at multiple levels.
For this part, we will recall some notations:
the prompt encoding $\bt_{ik} = \Phi_T(t_{ik})$;
$\bbf_{ik}^j$ is the frame encoding before VC;
$\hat{\be}_{ik}$ is the event encoding after VC; and
$\hat{\bv}_i$ the video encoding after VC.
We also consider a prompt representation for the full video obtained by mean pooling over all event-level prompts,
$\bt_i = \mean_k (\bt_{ik})$.

All losses below follow the contrastive loss (see  \cref{eq:nce}):

(i)~\textbf{CLIP-Event} applies a loss on the event representation obtained by mean pooling raw CLIP frame encodings:
$L^\text{CLIP}_\text{event} = L ( \mean_j(\bbf_{ik}^j), \bt_{ik} )$.

(ii)~\textbf{CLIP-Video} applies a loss on the video representation obtained by mean pooling raw CLIP frame encodings across the video and contrasting against the video-level prompt:
$L^\text{CLIP}_\text{video} = L ( \mean_{jk}(\bbf_{ik}^j), \bt_i )$.

(iii)~\textbf{VC-Event} applies a loss on the event representation post VC and the event-level prompt:
$L^\text{VC}_\text{event} = L ( \hat{\be}_{ik}, \bt_{ik} )$.
\cref{fig:model} (top-right) represents this loss.
Note how multiple events from the same video are used as negatives.

(iv)~\textbf{VC-Video}: applies a loss on the video representation post VC and video-level prompt:
$L^\text{VC}_\text{video} = L ( \hat{\bv}_i, \bt_i )$.
\cref{fig:model} (bottom-right) represents this loss.

We also use the symmetric version of the losses, \eg~$L(\bt_{ik}, \hat{\be}_{ik})$, which are not shown here for brevity.
We train the VC and adapt the backbone through a combination of all losses: $\ML = L^\text{CLIP}_\text{event} + L^\text{VC}_\text{event} + \lambda ( L^\text{CLIP}_\text{video} + L^\text{VC}_\text{video} )$.

\subsection{SRL Contains and Facilitates Hard Negatives}
\label{subsec:method:negatives} 

Contrastive learning requires hard negatives (HN) during training to prevent the model from finding the easy difference between the image and negative prompt. 

\paragraph{Natural hard negatives.}
Training with dense template prompts along with our batch creation strategy gives us natural hard negatives which promotes fine-grained, holistic, and efficient representation learning. 
For the CLIP-Event and VC-Event losses, negative prompts are obtained from different events of the same video which are generally quite similar (perhaps differing in only some of the verb/roles/nouns).
\cref{fig:model} top-right shows these natural HNs in light yellow background.
For example, a template prompt for a different event of the same video as described above (walking example):
\begin{displayquote}
\cgray{In this photo, the}
\emph{action} \cgray{is}
\underline{sit}
\cgray{where, the}
\emph{thing sitting} \cgray{is}
\underline{man with short} \underline{hair wearing collared shirt},
\emph{manner} \cgray{is}
\underline{casually}, and
\emph{scene} \cgray{of the event is}
\underline{apartment}.
\end{displayquote}

The subtle differences in verbs, roles, or their corresponding nouns, presents a challenging learning scenario for the model, resulting in multi-granularity representations.

\paragraph{Artificial hard negatives by replacing verb-role pairs.}
Although natural HNs provide challenging samples to the model,
template-based prompts can be used easily to create artificial HNs by replacing verb-role pairs.
Starting from the positive prompt, we replace the verb with a randomly sampled verb from the batch.
We also replace its corresponding roles, but keep the nouns unchanged.
Note, common roles such as \emph{direction}, \emph{manner}, \emph{scene} remain unchanged; making the prompts quite hard.
This allows the model to focus more on the \textit{action}.
We use $\MN_{vr}$ such negatives.
Differences to our walking example are in red:
\begin{displayquote}
\cgray{In this photo, the}
\emph{action} \cgray{is}
\underline{\cdeepred{jog}}
\cgray{where, the}
\emph{\cdeepred{jogger}} \cgray{is}
\underline{man with short hair} \underline{wearing collared shirt},
\emph{direction} \cgray{is}
\underline{forward}, and
\emph{scene} \cgray{of the event is}
\underline{apartment}.
\end{displayquote}

\paragraph{Incorporating negatives in the loss.}
HNs are only added to the event-level losses (VC-Event and CLIP-Event).
The loss function in \cref{eq:nce} is extended by including similarity scores between the visual information and the negative prompts in the denominator.

\subsection{Beyond VidSitu: Synthetic SRLs from Kinetics}
We advocate the use of dense text prompts for efficient post-pretraining of CLIP for videos.
Thus, while VidSitu is a natural fit, our work is \textit{not} specific to VidSitu.
While sparse labels generally perform worse (\eg~ViFi-CLIP),
we present an approach to leverage action labels in Kinetics-700 and create dense SRL prompts that are effective for adaptation.
Given video frames, we prompt LLaVA-1.6 ({\footnotesize \texttt{llava-hf/llava-v1.6-mistral-7b-hf}})
to generate role labels through questions: \eg~for the action label \emph{driving}, we ask ``who is driving?'' or ``what are they driving?''.
Thus, we create \textit{K-SRL}: 40k Kinetics videos with sparse action labels, augmented with dense and automatic (albeit noisy) SRL annotations.
Unlike VidSitu, these are single event descriptions.
Optionally, we can train \modelname{} with this data. 

\subsection{Implementation Details}
\label{subsec:method:details}

We use the OpenAI CLIP implementation and its associated checkpoints~\citep{clip}, restricting experiments to \texttt{ViT-B/32}, \texttt{ViT-B/16} and \texttt{ViT-L/14} models.
We add LoRA adapters to the CLIP image encoder and freeze both the CLIP image and text encoders.
We find $r{=}64$ rank to work well in our experiments.
Our VC module consists of 6 Tx encoder layers.
We use $\lambda{=}0.25$ for combining video- and event-level losses.
The number of artificial hard negatives $\MN_{vr}{=}4$.
We use a learning rate of $10^{-6}$ and the AdamW~\citep{adamw} optimizer.
Each video in VidSitu has $P{=}5$ events, and we sub-sample $T{=}4$ frames from each event, for a total of $L{=}20$ frames for a \SI{10}{\second} video.
We post-pretrain for 40 epochs on \textit{one 12GB RTX2080 GPU} for \texttt{ViT-B/32} \& \texttt{ViT-B/16}  and \textit{one 48GB RTX A6000 GPU} for \texttt{ViT-L/14} with a batch size of $B{=}20$ videos (100 event-text pairs).
\section{Experiments}
\label{sec:experiments}

We evaluate \modelname{} on a variety of video understanding tasks that require different levels of perceptual granularity.
Then, we present thorough ablations on the VidSitu dataset, providing insights about various design choices.

\subsection{Zero-shot Text-to-Video (T2V) Retrieval}
\label{subsec:exp:t2v}

\begin{table}[t!]
\footnotesize
\tabcolsep=0.07cm
\centering
\caption{Zero-shot text-to-video retrieval on LSMDC and MSRVTT.
Metrics are recall at 5 (R5$\uparrow$), at 10 (R10$\uparrow$), and median rank (MdR$\downarrow$).
Models in section B inherit CLIP, pretrained on 400M samples~\cite{clip}, and are further post-pretrained:
CLIP4Clip on HowTo100M-380k~\cite{clip4clip,howto100m};
ViFi on Kinetics-400~\cite{kinetics400};
and \modelname{} on VidSitu's 23k videos~\cite{vidsitu}.
See \cref{subsec:exp:t2v} for details.
}
\vspace{-2mm}
\label{table:zeroshot_t2v_main}
\begin{tabular}{l c c ccc ccc}
\toprule
& & & \multicolumn{3}{c}{LSMDC} & \multicolumn{3}{c}{MSRVTT}\\
\cmidrule(lr){4-6} \cmidrule(lr){7-9}
Method & DSize & Params & R5 & R10 & MdR & R5 & R10 & MdR \\
\midrule
\multicolumn{9}{l}{\emph{A. Non-CLIP based models}}\\
\addlinespace[2pt]
VideoCLIP~\cite{videoclip}       & 1M   & --    & --   & --   & --  & 22.2 & 30.0 & --  \\
Frozen~\cite{webvid2.5m}         & 5M   & 232M  & --   & --   & --  & 44.6 & 56.6 & 7   \\
Clover~\cite{clover}             & 5M   & --    & 29.2 & 38.2 & 24  & 49.5 & 60.0 & 6   \\
Singularity~\cite{singularity}   & 5M   & 209M  & --   & --   & --  & 50.2 & 59.5 & --  \\
HiTeA~\cite{hitea}               & 5M   & 297M  & 31.1 & 39.8 & --  & 54.2 & 62.9 & --  \\
ALPRO~\cite{alpro}               & 5.5M & 231M  & --   & --   & --  & 44.7 & 55.4 & --  \\
OmniVL~\cite{omnivl}             & 18M  & --    & --   & --   & --  & 58.4 & 66.6 & --  \\
VideoCoCa~\cite{videococa}       & 100M & 2.1B  & --   & --   & --  & 57.8 & 67.0 & --  \\
VIOLET~\cite{violet}             & 183M & 198M  & --   & --   & --  & 49.5 & 59.7 & --  \\
Florence~\cite{florence}         & 900M & 637M  & --   & --   & --  & 63.8 & 72.6 & --  \\
\midrule
\multicolumn{9}{l}{\emph{B. CLIP based models with simple post-pretraining}}\\
\addlinespace[2pt]
CLIP {\tiny ViT-B/32}                          & --           & 150M & 28.9 & 35.7 & 31  & 53.2 & 63.0 & 4 \\
CLIP {\tiny ViT-B/16}                          & --           & 150M & 32.4 & 40.4 & 21  & 55.0 & 65.3 & 4 \\
CLIP {\tiny ViT-L/14}                          & --           & 400M & 36.8 & 43.7 & 19  & 59.5 & 69.6 & \textbf{3} \\
CLIP4Clip~\cite{clip4clip}                            & 0.4M         & 150M & 28.5 & 36.4 & 28  & 57.0 & 66.9 & 4 \\
ViFi~\cite{vificlip}  {\tiny ViT-B/16}        & 0.3M         & 150M & 10.6 & 14.8 & 199 & 26.6 & 33.4 & 41 \\
ViFi-IFT~\cite{vificlip}  {\tiny ViT-B/16}    & 0.3M         & 150M & 32.6 & 39.9 & 25  & 57.6 & 67.1 & \textbf{3} \\
\addlinespace[2pt]
\rowcolor{LightCyan}
\modelname{} {\tiny ViT-B/32} & \textbf{23k} & 150M & 31.1 & 39.2 & 24           & 58.2 & 69.4 & \textbf{3} \\
\rowcolor{LightCyan}
\modelname{} {\tiny ViT-B/16} & \textbf{23k} & 150M & 35.7 & 43.8 & 17           & 59.7 & 71.8 & \textbf{3} \\
\rowcolor{LightCyan}
\modelname{} {\tiny ViT-L/14} & \textbf{23k} & 400M & \textbf{41.7} & \textbf{48.7} & \textbf{12} & 64.9 & 73.6 & \textbf{3} \\
\midrule
\multicolumn{9}{l}{\emph{C. CLIP based models with sophisticated architecture/modality extensions}}\\
\addlinespace[2pt]
BT-Adapter~\cite{bt_adapter}    & 2M    & 450M & 35.9 & 45.0 & --  & 64.7 & 73.5 & --  \\
ImageBind~\cite{imagebind}       & 3M   & 1B   & --   & --   & --  & 61.8 & 70.0 & --  \\
UMT-L~\cite{umtl}               & 5M   & 304M & 37.2 & 43.7 & --  & 58.1 & 66.7 & --  \\
TVTSv2~\cite{tvtsv2}            & 8.5M  & 1B   & 32.5 & 41.4 & 20  & 62.4 & 73.2 & \textbf{3} \\
VAST~\cite{vast}                 & 154M & 1.3B & --   & --   & --  & \textbf{68.3} & \textbf{73.9} & -- \\
\bottomrule
\end{tabular}
\vspace{-4mm}
\end{table}

We present results on zero-shot T2V retrieval.
To evaluate \modelname's representations for coarse video understanding, the VC is ignored here and retrieval scoring is performed by extracting frame-level features from \modelname{}'s backbone, followed by simple mean pooling, similar to~\citep{vificlip}.

We evaluate on two popular datasets: MSRVTT~\citep{msrvtt} and LSMDC~\citep{lsmdc}.
Results obtained using standard retrieval metrics (recall and mean/median rank) are presented in \cref{table:zeroshot_t2v_main}.
We also report post-pretraining efficiency by showing the dataset size and model size.
We categorize and describe the competing methods in three sections:

\paragraph{A. Non-CLIP based models}
generally require large models and large datasets to perform on par with CLIP based models.
Methods like VideoCoca~\citep{videococa} and Florence~\citep{florence} are trained on 100M and 900M video samples and adopt ViT-H as a backbone, resulting in 2.1B and 637M parameters respectively.
They require pretraining from scratch on multiple A100 GPUs for several days.

On the other hand, we build on top of CLIP (\texttt{ViT-B/16} with 150M parameters and \texttt{ViT-L/14} with 400M parameters), pretrained on 400M \textit{images} (an order of magnitude smaller when compared to the number of \textit{video frames} used above).
Our post-pretraining (LoRA adaptation) on 23k video clips from VidSitu is highly efficient: it can be performed on a single 12GB RTX 2080 GPU within 5 hours for the base model, and on a 48GB A6000 GPU within 18 hours for the large model.
On MSRVTT, our ViT-L/14 model outperforms VideoCoCa by 6.6\% R@10 and surpasses Florence by 1.0\% R@10 and outperforms all other models in this category (\cref{table:zeroshot_t2v_main}A).

\begin{figure*}[t]
\centering
\includegraphics[width=0.9\linewidth]{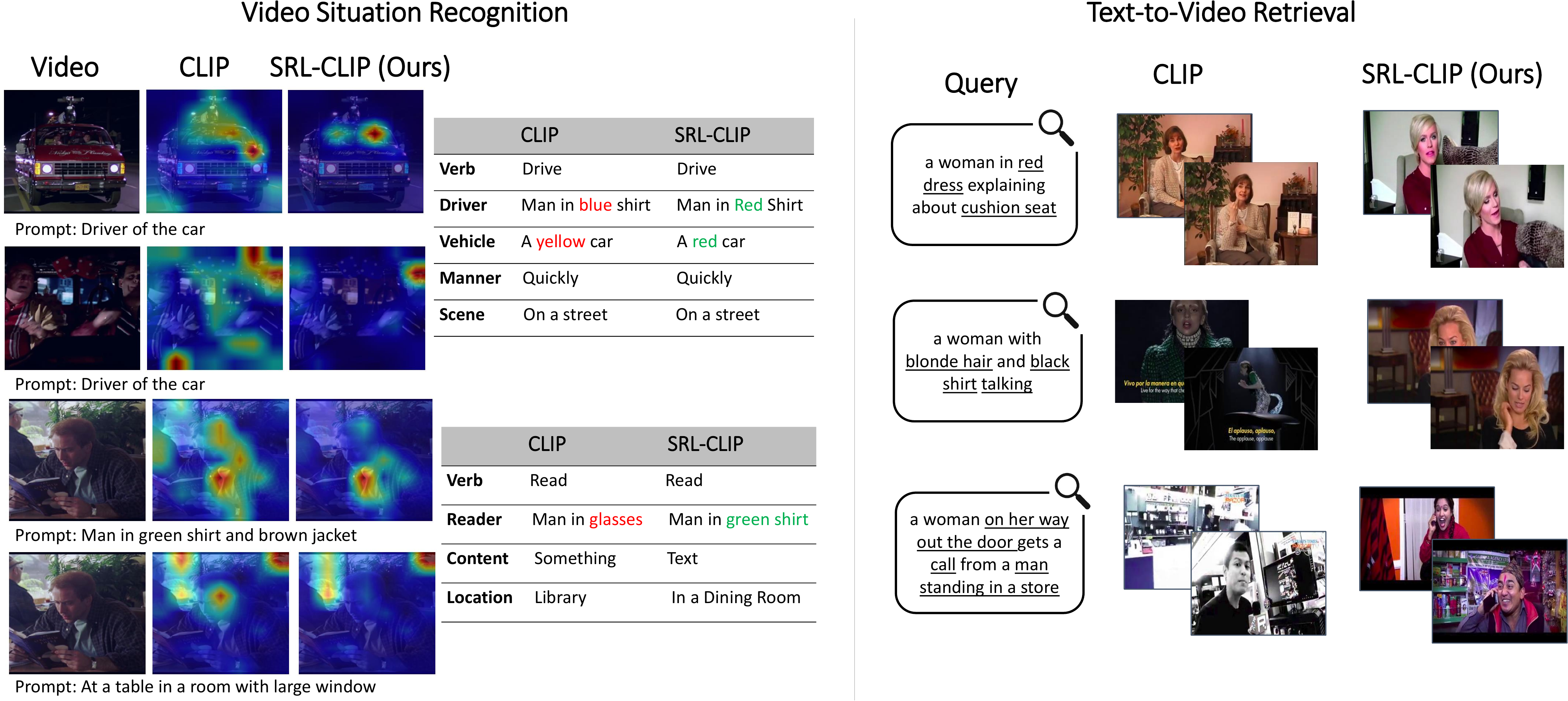}
\vspace{-3mm}
\caption{Qualitative results comparing CLIP and \modelname.
\textbf{VidSitu} (left) shows improved attention maps resulting in better noun (SRL) captions; and
\textbf{T2V Retrieval} (right) shows that \modelname{} has better awareness to details.}
\vspace{-4mm}
\label{fig:main_paper_qualitative}
\end{figure*}

\paragraph{B. CLIP-based model with simple finetuning.}
Our approach, \modelname, belongs to this category.
CLIP4Clip uses a Transformer-based frame feature aggregator, and is post-pretrained on 0.4M videos while ViFi-IFT uses simple mean pooling, and is fine-tuned on 300k videos.
Our approach, \modelname{} (ViT-L/14), fine-tuned on 23k videos with simple mean pooling, improves over CLIP4Clip and ViFi-IFT on all metrics. 
\Eg~R@10 improves by 12.3\% and 8.8\% on LSMDC, and 6.7\% and 6.5\% on MSRVTT in comparison to CLIP4Clip and ViFi-IFT respectively. 
\cref{fig:main_paper_qualitative}  (right) shows some qualitative results.
We observe that \modelname{} is able to correctly understand fine-grained details such as red dress (example 1) or associate multi-person events across multiple shots (example 3) better than base CLIP.

\paragraph{C. CLIP-based models with sophisticated extensions.}
In this part, we compare against models with much larger sizes (\eg~BT-Adapter~\citep{bt_adapter} at 450M, VAST~\citep{vast} at 1.3B parameters) that are often trained with multiple modalities (\eg~ImageBind~\citep{imagebind}, VAST), with post-pretraining datasets $85{-}6700\times$ larger than our 23k videos.
\modelname{} (ViT-L/14) achieves strong results:
(i)~TVTSv2~\citep{tvtsv2} is worse on LSMDC (7.3\% R@10) and is also slightly worse on MSRVTT (0.4\% R@10).
(ii)~ImageBind~\citep{imagebind}, on MSRVTT, has worse R@10 by 3.6\% and worse R@5 by 3.1\%.
(iii)~UMT-L~\citep{umtl} is worse by 5.0\% R@10 on LSMDC, and is worse by 6.9\% R@10 on MSRVTT.
Finally, \modelname{} outperforms BT-Adapter by 0.1\% R@10 on MSRVTT.
VAST performs marginally better than \modelname{} (0.3\% R@10 on MSRVTT), however, both these models have higher model footprints (450M and 1.3B) and are post-pretrained on 2M and 154M videos (ours on 23k videos).

These results highlight the efficacy achievable through post-pretraining on dense and structured SRL, as opposed to narrations or captions.

\begin{table}[t]
\footnotesize
\tabcolsep=0.07cm
\caption{ZS T2V retrieval performance on MSRVTT and LSMDC on the \texttt{ViT-L/14} variants. VS = VidSitu, KS = Kinetics-SRL.}
\vspace{-2mm}
\centering
\begin{tabular}{ll ccc ccc ccc}
\toprule
& & & & & \multicolumn{3}{c}{MSRVTT} & \multicolumn{3}{c}{LSMDC}\\
\cmidrule(lr){6-8} \cmidrule(lr){9-11}
\# & Model & DSize & VS & KS & R@5 & R@10 & MdR & R@5 & R@10 & MdR\\
\midrule
1 & CLIP & -- & -- & -- 
  & 59.0 & 69.6 & 3 
  & 36.8 & 43.7 & 19.0 \\
2 & \modelname{} & 23k & \checkmark{} & -- 
  & 64.9 & 73.6 & 3 
  & \textbf{41.7} & 48.7 & 11.5 \\
3 & \modelname{} & 40k & -- & \checkmark{} 
  & 64.1 & 73.8 & 3 
  & 40.2 & 47.8 & 12.0 \\
\rowcolor{LightCyan}
4 & \modelname{} & 63k & \checkmark{} & \checkmark{} 
  & \textbf{65.0} & \textbf{74.0} & 3 
  & 41.2 & \textbf{50.8} & \textbf{10.0} \\
\bottomrule
\end{tabular}
\label{table:k700_vsitu}
\vspace{-2mm}
\end{table}

\paragraph{Kinetics-700.}
\cref{table:k700_vsitu} shows the results for adapting CLIP with K-SRL and VidSitu.
Even though the K-SRL annotations are noisy, we improve R@5 by 5.1\% for CLIP (row3 \vs~row1) for MSRVTT while 3.4\% for LSMDC.
Training on K-SRL followed by VidSitu further improves over only training on VidSitu.
Row4 \vs~row2 shows a 0.8 reduction in MnR and a 0.4\% increase in R@10 for MSRVTT.
For LSMDC, there is a significant reduction in MnR (by 6.2) and a 2.1\% increase in R@10.

\subsection{Holistic Video Understanding}
\begin{table}[t]
\small
\tabcolsep=0.12cm
\centering
\caption{Performance on VidSitu~\citep{vidsitu}.
Captioning performance measured through CIDEr.
We see a large performance improvement over base CLIP, while also achieving a new SoTA.
}
\vspace{-2mm}
\label{table:vidsitu_sota}
\begin{tabular}{l cc c}
\toprule
Method & Vb@1 $\uparrow$ & Vb@5 $\uparrow$ & CIDEr $\uparrow$ \\
\midrule
VidSitu~\cite{vidsitu} & 46.79 & 75.90 & 46.01\\
Slow-D+TxE+TxD~\cite{xiao2022hierarchical} & - & - & 60.34\\
VideoWhisperer~\cite{gvsr} & 45.06 & 75.59 & 68.54\\
\midrule
CLIP \scriptsize{(ViT B/32)}
        & 44.68 & 79.79 & 55.14 \\
\rowcolor{LightCyan}
\modelname{} \scriptsize{(ViT B/32)}
        & \textbf{45.88} & \textbf{80.66} & \textbf{71.50} \\
\midrule
CLIP \scriptsize{(ViT B/16)}
        & 45.83 & 80.12 & 54.25 \\
\rowcolor{LightCyan}
\modelname{} \scriptsize{(ViT B/16)}
        & \textbf{48.78} & \textbf{81.95} & \textbf{72.11} \\
\midrule
CLIP \scriptsize{(ViT L/14)}
        & 46.57 & 78.73 & 60.79 \\
\rowcolor{LightCyan}
\modelname{} \scriptsize{(ViT L/14)}
       & \textbf{52.36} & \textbf{83.91} & \textbf{76.24} \\
\bottomrule
\end{tabular}
\vspace{-3mm}
\end{table}

We evaluate \modelname's holistic video understanding ability from two perspectives:
(i)~VidSitu~\citep{vidsitu} requires fine-grained perception to generate complex structured outputs; and
(ii)~Dense video captioning~\citep{activitynet} requires localizing and describing actions in longer (few minute) videos.

\paragraph{Video situation recognition} (VidSitu)~\citep{vidsitu}
requires a model to predict the primary action verb in each event and generate noun captions for the verb-appropriate set of roles. 
We adopt VideoWhisperer's three-stage architecture~\citep{gvsr} as it is the SoTA on VidSitu and remove the 165 object (Faster-RCNN) features and 5 event/action (SlowFast) features.
Instead, as a simplification, we represent the $P{=}5$ events as mean-pooled CLIP encoded frames.

In Table~\ref{table:vidsitu_sota}, we see that CLIP features perform on par with previous best methods on verbs (action understanding) achieving close to 45\% Vb@1 accuracy.
However, they perform much worse on CIDEr (55\%) that evaluates quality of semantic role labels.
This is expected as SRL requires fine-grained understanding that CLIP lacks~\citep{lemons}.

Next, we evaluate \modelname{} by representing the $P$ event features with event-level contextualized tokens (post VC).
Verb accuracy improves by 5.8\% for ViT L/14, while we see a large increase in SRL prediction performance with CIDEr improving by 15.5\%.
We acknowledge that a part of this gain may be due to the in-domain post-pretraining on VidSitu, especially training VC from scratch.
Nevertheless, this demonstrates the effectiveness of our adaptation process and the video contextualizer that compresses information effectively.
With this, we also set a new SoTA on VidSitu.
Finally, in \cref{fig:main_paper_qualitative}, we show that our adapted model is better suited for fine-grained reasoning with focused attention maps and accurate SRL prediction.

\begin{table}[t]
\footnotesize
\tabcolsep=0.07cm
\centering
\caption{Performance on the recent VELOCITI benchmark. We report the ClassicVLE accuracy across 7 different tests on the \texttt{ViT-L/14} variant. VS = VidSitu, KS=Kinetics-SRL.}
\vspace{-2mm}
\begin{tabular}{lcccccccc}
\toprule
Method & Data & \shortstack{Ag\\Rand} & \shortstack{Ag\\Bind} & \shortstack{Act\\Adv} & \shortstack{Act\\Man} & \shortstack{Act\\Bind} & \shortstack{Ag\\Coref} & Chrono \\
\midrule
CLIP & -- & 82.47 & 56.07 & 64.84 & 54.80 & \textbf{58.48} & 53.06 & 49.43 \\
SRL-CLIP & VS & \textbf{87.97} & 55.17 & 63.93 & \textbf{60.26} & 55.01 & 55.16 & 48.70 \\
SRL-CLIP & KS & 85.80 & \textbf{56.27} & \textbf{65.75} & 56.77 & 57.52 & \textbf{57.82} & 48.54 \\
SRL-CLIP & VS+KS & 87.17 & 52.98 & 64.38 & 58.95 & 55.09 & 54.28 & \textbf{49.59} \\
\bottomrule
\end{tabular}
\vspace{-3mm}
\label{tab:velociti_main}
\end{table}

\paragraph{VELOCITI}~\citep{velociti} is a recent benchmark used to test compositional reasoning abilities in video-language models.
It has a series of 7 tests to evaluate different aspects of compositional reasoning for agents, actions, and event ordering. 
The results are shown in \cref{tab:velociti_main}.
\modelname{} outperforms Base CLIP in all tests except ActBind.
Further, \modelname{} trained on VidSitu attains the highest accuracy on AgRand and ActMan, whereas when it is trained on Kinetics, it performs well on AgBind, ActAdv and AgCoref.
Finally, \modelname{} trained on both VidSitu and Kinetics obtains slight improvements on the Chrono test.

\subsection{Ablations}
We evaluate the impact of various design choices, with results on VidSitu or MSRVTT.
All ablations are performed on the CLIP ViT L/14 model.

\paragraph{How to adapt CLIP?}
We start with identifying how and which layers of the backbone should be adapted.
\cref{table:arch_adapt} shows various options from full (F) or partial fine-tuning (P), and using LoRA modules (L) on the image encoder (IE) and text encoder (TE).
Column 2 \vs~3 and 4 \vs~5 show that freezing TE improves performance.
We suspect this is because it encourages IE to align with fine-grained descriptions.
Column 5 shows good SRL CIDEr performance with LoRA for IE and frozen TE.
Comparing columns 1 and 5 we see that full fine-tuning (F) is not only costly, but also performs worse than LoRA (1.1\% Vb@1 and 1\% CIDEr), likely due to concept forgetting.
We adopt LoRA fine-tuning for the IE with a frozen TE as our default configuration.

\begin{table}[t]
\footnotesize
\centering
\tabcolsep=0.15cm
\caption{Ablation on CLIP adaptation strategy. F: Full fine-tuning, P: Partial fine-tuning, L: LoRA, IE/TE: Image/Text encoder.}
\vspace{-2mm}
\label{table:arch_adapt}
\begin{tabular}{l c cccc >{\columncolor{LightCyan}}c}
\toprule
\multirow{2}{*}{Backbone} & IE & F & P & P & L & L \\
& TE & - & P & - & L & - \\
\midrule
\multirow{2}{*}{VidSitu} & Vb@1 & 43.53 & 42.83 & 42.87 & \textbf{45.53} & 44.65 \\
& CIDEr & 70.27 & 67.83 & 71.00 & 70.14 & \textbf{71.27} \\
\bottomrule
\end{tabular}
\vspace{-2mm}
\end{table}

\paragraph{Impact of loss functions}
is presented in \cref{table:losses}.
Directly adapting the CLIP backbone with the CLIP-Event and CLIP-Video losses (R1-2) results in poor SRL performance on VidSitu.
The benefits of a video contextualizer (VC), both during training and downstream evaluation, are seen in R4 that uses VC-Event and VC-Video losses, achieving good results on VidSitu SRL (R4: 75.0\% \vs~R2: 58.3\% CIDEr for \texttt{ViT-L/14}).
R5-6 combine both CLIP and VC losses.
We choose R6 as the default model as it achieves the highest geometric mean (good trade-off) between Vb@1 and CIDEr scores, yielding 52.36\% Vb@1 and 76.24 CIDEr.

\begin{figure}[t]
\centering
\captionof{table}{Ablation on loss functions. Results on VidSitu. CE: CLIP-Event, CV: CLIP-Video, VCE: VC-Event, VCV: VC-Video.}
\label{table:losses}
\vspace{-2mm}
\footnotesize
\tabcolsep=0.12cm 
\begin{tabular}{l cccc cccc}
\toprule
& \multicolumn{4}{c}{Losses} & \multicolumn{2}{c}{ViT-L/14} & \multicolumn{2}{c}{ViT-B/32} \\
\cmidrule(lr){2-5} \cmidrule(lr){6-7} \cmidrule(lr){8-9}
& CE & CV & VCE & VCV & Vb@1 & CIDEr & Vb@1 & CIDEr \\
\midrule
1 & \cmark & - & - & - & 52.78 & 59.08 & 44.77 & 58.37 \\
2 & \cmark & \cmark & - & - & 51.93 & 58.27 & \textbf{45.38} & 57.34 \\
3 & - & - & \cmark & - & \textbf{52.72} & 75.21 & 43.75 & 71.19 \\
4 & - & - & \cmark & \cmark & 52.08 & 75.03 & 45.12 & 70.02 \\
5 & \cmark & - & \cmark & \cmark & 52.66 & 73.66 & 44.46 & \textbf{72.10} \\
\rowcolor{LightCyan}
6 & \cmark & \cmark & \cmark & \cmark & 52.36 & \textbf{76.24} & 44.65 & 71.27 \\
\bottomrule
\end{tabular}
\vspace{-2mm}
\end{figure}

\begin{table}[t]
\footnotesize
\centering
\caption{Ablation study on when should VC be used. PPT: Post-pretraining, DT: downstream-task.}
\label{table:vc}
\vspace{-2mm}

\textbf{(a) VidSitu (Video Situation Recognition)}
\vspace{2pt}

\begin{tabular}{lcc cccc}
\toprule
& \multicolumn{2}{c}{VC} & \multicolumn{2}{c}{ViT-B/32} & \multicolumn{2}{c}{ViT-L/14} \\
\cmidrule(lr){2-3} \cmidrule(lr){4-5} \cmidrule(lr){6-7}
  & PPT & DT & Vb@1 & CIDEr & Vb@1 & CIDEr \\
\midrule
1 & \multicolumn{2}{c}{CLIP} & 44.7 & 55.1 & 46.6 & 60.8 \\
2 & - & - & 45.4 & 57.3 & 51.9 & 58.3 \\
3 & \cmark & - & \textbf{46.7} & 59.8 & \textbf{53.1} & 63.4 \\
4 & \cmark & \cmark & 45.9 & \textbf{71.5} & 52.4 & \textbf{76.2} \\
\bottomrule
\end{tabular}

\vspace{6pt}

\textbf{(b) MSRVTT (zero-shot T2V retrieval)}
\vspace{2pt}
\tabcolsep=0.08cm
\begin{tabular}{lcc cccc cccc}
\toprule
& \multicolumn{2}{c}{VC} & \multicolumn{4}{c}{ViT-B/32} & \multicolumn{4}{c}{ViT-L/14} \\
\cmidrule(lr){2-3} \cmidrule(lr){4-7} \cmidrule(lr){8-11}
  & PPT & DT & R@5 & R@10 & Mn.R & Md.R & R@5 & R@10 & Mn.R & Md.R \\
\midrule
1 & \multicolumn{2}{c}{CLIP} & 53.2 & 63.0 & 41.2 & 4 & 59.0 & 69.6 & 33.6 & 3 \\
2 & - & - & 57.5 & 68.1 & 29.5 & 4 & 64.3 & \textbf{75.7} & \textbf{21.7} & \textbf{2} \\
3 & \cmark & - & \textbf{58.2} & \textbf{69.4} & \textbf{27.8} & \textbf{3} & \textbf{64.7} & 73.8 & 22.5 & 3 \\
4 & \cmark & \cmark & 15.9 & 24.7 & 120.2 & 46 & 32.8 & 43.5 & 71.4 & 15 \\
\bottomrule
\end{tabular}
\vspace{-3mm}
\end{table}

\paragraph{When should VC be used?}
We investigate the need for VC and its influence on downstream results in \cref{table:vc}.
We evaluate on two tasks, in-domain VidSitu and zero-shot T2V retrieval on MSRVTT.

First, row 1 (R1) shows results for the original CLIP model without adaptation, copied here for ease of comparison.
In R2, we directly use CLIP-Video and CLIP-Event losses; while R3 uses VC only for post-pretraining (PPT).
In all three cases, downstream tasks (DT) are performed by extracting features directly from the CLIP backbones.

For ViT-B/32, we observe improvements across all metrics as we step from R1 to R2 to R3.
For zero-shot T2V on MSRVTT, we see that R2 (direct CLIP PPT) brings a large improvement over R1 (5.1\% R@10).
This highlights the benefit of using structured SRL for efficient CLIP video adaptation.
Nevertheless, including VC results in a further 1.3\% improvement on R@10 (R3).
On VidSitu, we see steady improvements on CIDEr: 2.2\% from R1 to R2, and 2.5\% from R2 to R3.

Next, we analyze what happens when using VC in both the post-pretraining and the downstream task (R4), again for ViT-B/32.
Specifically, we use the $P$ contextualized event representations $\hat{\be}_{ik}$ for VidSitu and the single contextualized video representation $\hat{\bv}_i$ for MSRVTT.
The VC helps compress and contextualize multiple events in a video, leading to strong performance improvements on VidSitu (SRL CIDEr improves by 11.7\% over R3).
As VC is trained only on VidSitu, it learns to accommodate the complexity of VidSitu.
For example, the overall video representation is trained to match against a dense description of multiple events.
However, when VC is used directly on MSRVTT, the performance collapses as the captions in MSRVTT are not as descriptive.

We observe similar overarching trends for the larger \texttt{ViT-L/14} model, where VC significantly boosts fine-grained VidSitu performance (a 12.8\% CIDEr improvement in R4 over R3), though for coarse retrieval on MSRVTT, direct post-pretraining without VC (R2) is sufficient to achieve peak performance.

\section{Conclusion}
\label{sec:conclusion}

We proposed a recipe for efficient and effective post-pretraining of CLIP to adapt it to videos, using a densely annotated video-caption dataset.
Specifically, we curated template-based prompts from a video SRL dataset~\citep{vidsitu}, that captures visual concepts in a video holistically using action verbs, person-object nouns, attributes, locations, \etc.
We hypothesized that post-pretraining on dense prompts provides a strong learning signal to the visual encoder, which not only allows for highly efficient but also effective adaptation to videos.
Our adapted model SRL-CLIP, exhibited superior or comparable performance on zero-shot text-to-video retrieval against SOTA models having $4{-}8\times$ more parameters and post-pretrained on up to $6000\times$ more data.
Additionally, SRL-CLIP shows good downstream generalization, consistently outperforming CLIP on multiple video benchmarks requiring different levels of perceptual granularity.
Specifically, we achieve SoTA performance on a fine-grained and holistic video understanding benchmark VidSitu~\citep{vidsitu}.

\paragraph{Acknowledgments.}
The project was supported by funding from
SERB SRG/2023/002544.
MT also thanks support from the Google India Faculty Award.

\newpage

{
    \small
    \bibliographystyle{ieeenat_fullname}
    \bibliography{bib/longstrings, bib/refs}
}

\setcounter{page}{1}
\maketitlesupplementary

\section*{Appendix}
We present additional results and discussions in the supplementary material.
\cref{sec:supp_ablations} starts with additional ablation experiments, focusing on hard negatives (HNs) and the low-rank adaptation (LoRA) module.
In \cref{sec:supp_qualitative} we present qualitative results on the various tasks shown in the main paper.
This provides an opportunity to better understand the working of \modelname. All ablation studies are conducted using the \texttt{ViT-L/14} variant. Finally, in \cref{sec:limitations} we discuss some of the limitations of our work.

\begin{table}[t]
\centering
\caption{Impact of varying the no. of verb-role hard negatives, $\MN_{vr}$.}
\label{table:verb_hn}
\vspace{1mm}
\tabcolsep=0.4cm
\begin{tabular}{c cc}
\toprule
& \multicolumn{2}{c}{VidSitu}\\
\cmidrule(lr){2-3}
$\MN_{vr}$ & Vb@1 & CIDEr \\
\midrule
0 & 52.38 & 75.63\\
1 & 51.72 & 74.78\\
2 & 52.54 & 74.02\\
3 & 53.26 & 74.88\\
4 & \textbf{52.36} & \textbf{76.24}\\
\bottomrule
\end{tabular}
\end{table}

\begin{table}[t]
\centering
\caption{Impact of adapting different weights using LoRA. q, k, v, and o are the query, key, value, and output projection matrices in the self-attention block. fc and proj are the two MLPs after the self-attention module.}
\label{table:lora_params}
\vspace{1mm}
\tabcolsep=0.4cm
\begin{tabular}{l cc}
\toprule
& \multicolumn{2}{c}{VidSitu}\\
\cmidrule(lr){2-3}
Weight type & Vb@1 & CIDEr\\
\midrule
q k v &  \textbf{52.36} & \textbf{76.24}\\
q k v o & \textbf{52.62} & 73.13\\
q k v o fc & 51.22 & 73.61\\
q k v o fc proj & 51.83 & 71.86\\
\bottomrule
\end{tabular}
\end{table}

\section{Additional Ablations}
\label{sec:supp_ablations}

\paragraph{How many Hard Negatives to use?}
\cref{table:verb_hn} shows the impact of varying the number of hard negatives (HNs).
Here, we see that as $\MN_{vr}$ increases, the verb prediction accuracy increases while CIDEr drops by a small amount.
This is expected as verb-role HNs tend to improve verb prediction performance.
The best performance is considered at the geometric mean between Vb@1 and CIDEr on the VidSitu task.

\paragraph{What is the best LoRA configuration?}
In \cref{table:lora_params}, we study the impact of adapting different weights of the \modelname{} image encoder with LoRA.
We get the best performance when we include LoRA modules only for the attention weights in the Transformer (``q k v'' -- $W_q$, $W_k$, $W_v$) while keeping everything else frozen. Hence, in our default model, we only adapt the self-attention matrices (parameters) with LoRA.  

\begin{table}[t]
\centering
\caption{ZS T2V on MSRVTT with CLIP {\scriptsize ViT B/32}. We increase the amount of VidSitu training data, resulting in a consistent improvement in performance for R@10 and MnR.}
\label{table:vsitu_scale}
\vspace{1mm}
\tabcolsep=0.4cm
\begin{tabular}{l cc}
\toprule
& \multicolumn{2}{c}{VidSitu}\\
\cmidrule(lr){2-3}
Data & Vb@1 & CIDER \\
\midrule
0\% & 46.57 & 60.79\\
50\% & 50.82 & 70.72\\
70\% & 50.62 & 73.31\\
90\% & 51.36 & 73.55\\
100\% & \textbf{52.36} & \textbf{76.24} \\
\bottomrule
\end{tabular}
\end{table}

\paragraph{Training on subsets of VidSitu.}
The results of increasing the post-pretraining dataset from 10\% to 100\% are presented in \cref{table:vsitu_scale}.
The table shows a consistent decline in mean rank and an improvement in R@10 with additional data.
Additionally, with just 10\% of the data, a mere 2,300 videos, we see a 3.9\% boost in R@5 and a 2.6\% in R@10; indicating the effectiveness of dense SRL prompts.

\paragraph{Detailed descriptions incur labeling costs, but save on compute costs.}
The use of dense prompts results in efficient training that completes in 5 hours on a \emph{single} RTX2080 GPU (\SI{12}{\giga\byte}).
In contrast, large models adapted on large (noisy) datasets incur significantly higher compute costs requiring many and larger GPUs.
\Eg~BT-Adapter~\citep{bt_adapter} uses 8 V100 GPUs,
TVTSv2~\citep{tvtsv2} uses 80 V100 GPUs, and
VAST~\citep{vast} uses 64 V100 GPUs.
Considering the typical number of hyperparameter evaluations and experiments required, improving data quality is not only cheaper but also more viable in the long-term.

\begin{table*}[t]
\centering
\small
\tabcolsep=0.15cm
\begin{tabular}{p{3.6cm} p{3.6cm} p{3.6cm}}
\toprule
\multicolumn{1}{c}{Positive Prompt}
&
\multicolumn{1}{c}{Natural Hard Negatives}
&
\multicolumn{1}{c}{Verb-role Hard Negatives}
\\
\midrule
\cgray{In this photo, the}
\emph{action} \cgray{is}
\cblue{speak}
\cgray{where, the}
\emph{talker} \cgray{is}
\cblue{man standing in yellow sweatshirt}, \cgray{the}
\emph{hearer} \cgray{is}
\cblue{woman with scarf}, \cgray{the}
\emph{manner} \cgray{is}
\cblue{standing in the middle of a full airplane}, \cgray{the}
\emph{scene} \cgray{of the event is}
\cblue{an airplane}.
&
\cgray{In this photo, the}
\emph{action} \cgray{is}
\cdeepred{turn}
\cgray{where, the}
\emph{\cdeepred{the turner}} \cgray{is}
\cblue{man standing in yellow sweatshirt}, \cgray{the}
\emph{\cdeepred{the thing turning}} \cgray{is}
\cdeepred{his body}, \cgray{the}
\emph{\cdeepred{direction}} \cgray{is}
\cblue{towards woman with scarf}, \cgray{the}
\emph{scene} \cgray{of the event is}
\cblue{an airplane}.
&
\cgray{In this photo, the}
\emph{action} \cgray{is}
\cblue{\cdeepred{look}}
\cgray{where, the}
\emph{\cdeepred{looker}} \cgray{is}
\cblue{man standing in yellow sweatshirt}, \cgray{the}
\emph{\cdeepred{thing looked at}} \cgray{is}
\cblue{woman with scarf}, \cgray{the}
\emph{\cdeepred{direction}} \cgray{is}
\cdeepred{\emph{is to his back}}, 
\emph{manner} \cgray{is}
\cblue{standing in the middle of a full airplane}, \cgray{the}
\emph{scene} \cgray{of the event is}
\cblue{an airplane}.
\\
\midrule
\cgray{In this photo, the}
\emph{action} \cgray{is}
\cblue{open}
\cgray{where, the}
\emph{opener} \cgray{is}
\cblue{man in brown jacket and man in gray suit}, \cgray{the}
\emph{the thing opening} \cgray{is}
\cblue{trunk of taxi}, \cgray{the}
\emph{manner} \cgray{is}
\cblue{annoyed}, \cgray{the}
\emph{scene} \cgray{of the event is}
\cblue{near a taxi}.
&
\cgray{In this photo, the}
\emph{action} \cgray{is}
\cdeepred{hoist}
\cgray{where, the}
\emph{\cdeepred{lifter}} \cgray{is}
\cblue{man in brown jacket and man in gray suit}, \cgray{the}
\emph{\cdeepred{thing going up}} \cgray{is}
\cdeepred{dead body}, \cgray{the}
\emph{\cdeepred{direction}} \cgray{is}
\cdeepred{up into trunk of taxi}, \cgray{the}
\emph{scene} \cgray{of the event is}
\cblue{near a taxi}.
&
\cgray{In this photo, the}
\emph{action} \cgray{is}
\cblue{\cdeepred{respond}}
\cgray{where, the}
\emph{\cdeepred{replier}} \cgray{is}
\cblue{man in brown jacket and man in gray suit}, \cgray{the}
\emph{scene} \cgray{of the event is}
\cblue{near a taxi}.
\\
\midrule
\cgray{In this photo, the}
\emph{action} \cgray{is}
\cblue{bow}
\cgray{where, the}
\emph{bower} \cgray{is the}
\cblue{woman in glasses}, \cgray{the}
\emph{bowed to} \cgray{is}
\cblue{man wearing black}, \cgray{the}
\emph{manner} \cgray{is}
\cblue{on her knees}, \cgray{the}
\emph{scene} \cgray{of the event is}
\cblue{in a well lit room}.
&
\cgray{In this photo, the}
\emph{action} \cgray{is}
\cdeepred{photograph, take a picture}
\cgray{where, the}
\emph{\cdeepred{photographer}} \cgray{is the}
\cblue{man in black suit}, \cgray{the}
\emph{\cdeepred{subject}} \cgray{is}
\cblue{woman in glasses},
\emph{scene} \cgray{of the event is}
\cblue{in a well lit room}.
&
\cgray{In this photo, the}
\emph{action} \cgray{is}
\cblue{\cdeepred{smash}}
\cgray{where, the}
\emph{\cdeepred{smasher}} \cgray{is the}
\cblue{woman in glasses}, \cgray{the}
\emph{\cdeepred{smashed}} \cgray{is}
\cblue{man wearing black}, \cgray{the}
\emph{\cdeepred{direction}} \cgray{is}
\cblue{\cdeepred{on patients face}}, \cgray{the}
\emph{scene} \cgray{of the event is}
\cblue{in a well lit room}.
\\
\bottomrule
\end{tabular}
\caption{We show the naturally occurring hard negatives in a batch as well as the process of converting a standard positive prompt into hard negatives by swapping verb-role information.
The template is shown in \cgray{gray}, \eg~\cgray{In this photo,}.
The action and roles are shown in \emph{italics}, \eg~\emph{action, talker, hearer}.
The correct prompt values (verbs or nouns) are in \cblue{cobalt blue}, \eg~\emph{speak, man standing in yellow sweatshirt}; and
the replaced verbs, roles, or nouns are in \cdeepred{deep red}.
We swap the verb and roles in verb-role hard negatives while keeping the same nouns and performing some mapping between previous and new roles.
}
\label{tab:supp_hn_examples}
\end{table*}

\paragraph{More examples of \emph{natural} and \emph{artificial} hard negatives}
are shown in \cref{tab:supp_hn_examples}.
Notice how the naturally occurring hard negatives are good enough to learn strong video representations even without the need for artificial hard negatives.
Different from most works that only use text-based negatives, VidSitu also facilitates visual natural negatives for the same text.
Note how the hard negative captions are very plausible; in example 1 the action \textit{look} instead of \textit{speak}.

\begin{figure*}[t]
\centering
\includegraphics[width=\linewidth]{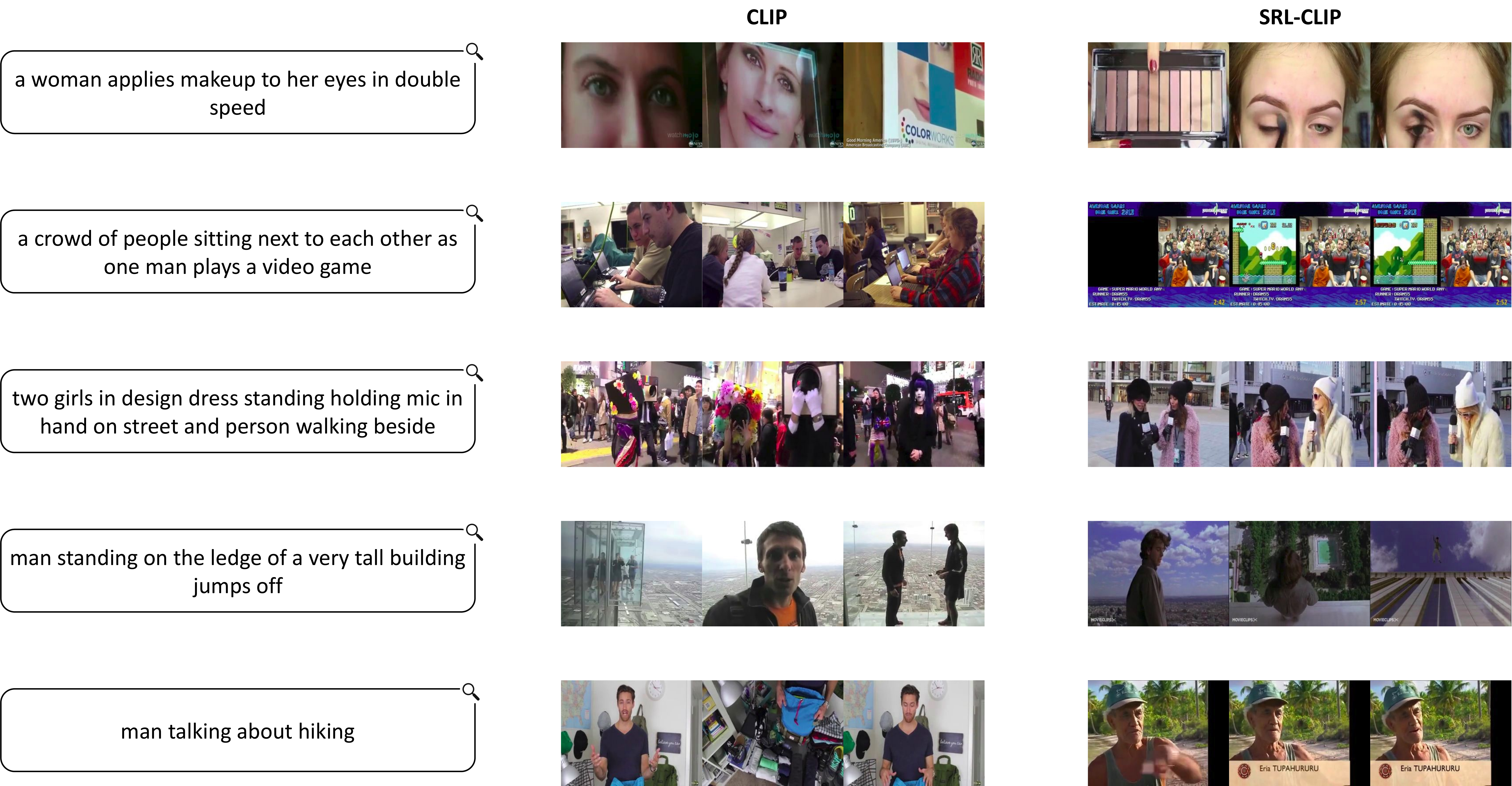}
\vspace{-5mm}
\caption{
Zero-shot text-to-video retrieval on the MSRVTT dataset. We show three frames of the top-1 retrieved video for each query. We can see that \modelname{} outperforms CLIP, specially when compositional reasoning is required. The last row shows a failure case. Although \modelname{} retrieves a video in which a man is talking, and potentially with more appropriate background, he is not talking about hiking.
}
\vspace{-2mm}
\label{fig:msrvtt_qual}
\end{figure*}

\begin{figure*}[t]
\centering
\includegraphics[width=\linewidth]{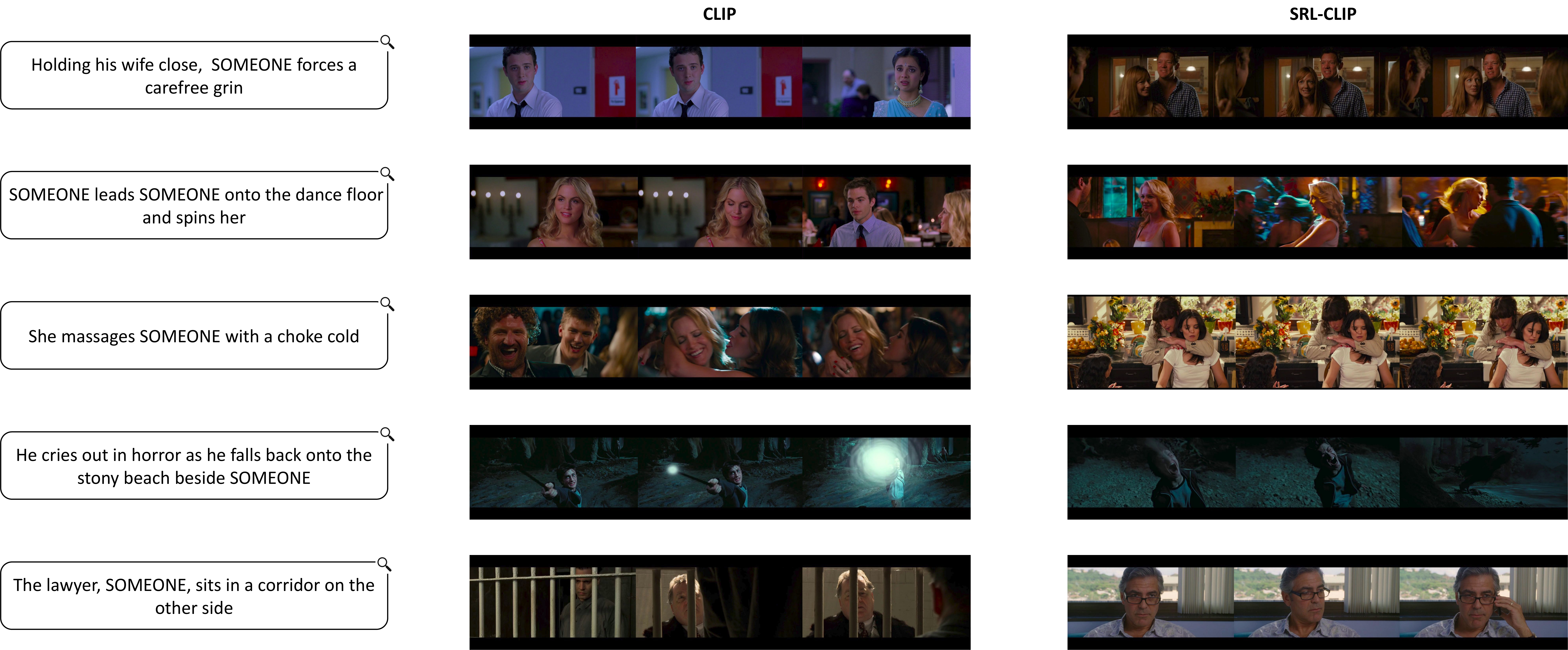}
\vspace{-5mm}
\caption{
Zero-shot text-to-video retrieval on the LSMDC dataset. We show three frames of the top-1 retrieved video for each query. We can again notice that \modelname{} performs better than CLIP when compositional reasoning is needed. The last row shows a failure case.
}
\vspace{-3mm}
\label{fig:lsmdc_qual}
\end{figure*}

\section{Qualitative Results}
\label{sec:supp_qualitative}

We now present qualitative results on 3 datasets.
When not mentioned otherwise, we use the default variant of \modelname.

\paragraph{MSRVTT.}
We show zero-shot text-to-video retrieval on the MSRVTT dataset in \cref{fig:msrvtt_qual}.
We can see that \modelname{} performs much better than CLIP when the queries have a compositional nature.
The last row shows a failure case. Although \modelname{} retrieves a video in which a man is talking, he is not talking about hiking.
It is hard to pick the right video just by using the visual modality, as hiking is not very clear by just watching the video.
In fact, \modelname{} retrieves a video shot outdoors, which may have be associated with hiking rather than the indoor video.

\paragraph{LSMDC.}
We show zero-shot text-to-video retrieval on the LSMDC dataset in \cref{fig:lsmdc_qual}.
LSMDC is a much harder dataset compared to MSRVTT, as it is based on movies that contain more dynamic shot changes.
Also, the agent/patient of an action is annotated as ``SOMEONE", unlike VidSitu, where they are described according to their characteristics, making it even more challenging.
We can see that \modelname{} outperforms CLIP here as well.

\begin{figure*}[t]
\centering
\includegraphics[width=\linewidth]{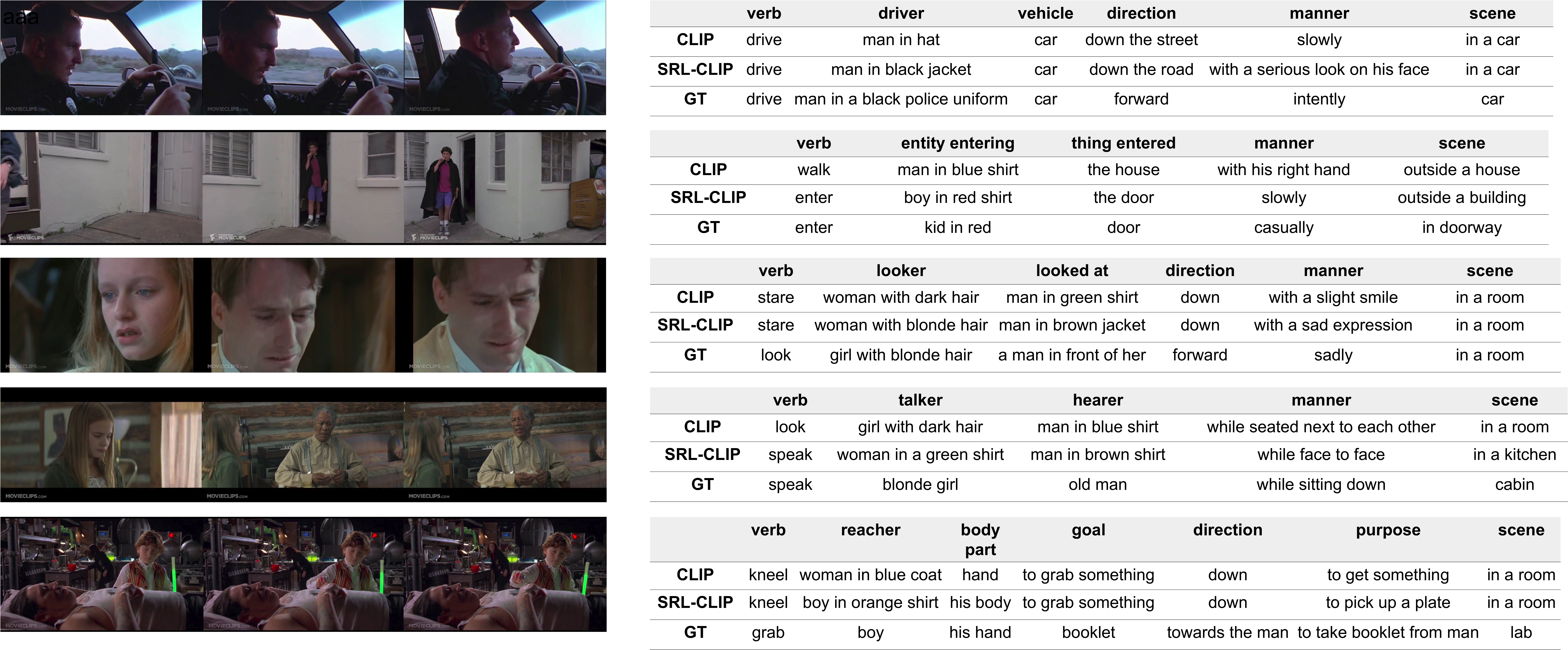}
\caption{
Video Situation Recognition on 5 videos. \modelname{} performs much better than CLIP in picking the right attribute of an entity. The last row shows a failure case where the semantic role labels predicted by \modelname{} deviates from the ground-truth (GT).
}
\label{fig:vidsitu_qual}
\end{figure*}

\paragraph{VidSitu.}
\cref{fig:vidsitu_qual} shows the qualitative results on video situation recognition for 5 videos. 
\modelname{} outperforms CLIP, especially when picking attributes like color.
\modelname{} is also better at predicting the role \emph{manner} (which captures the expression/emotion of the person), which CLIP struggles with.
However, both \modelname{} and CLIP show similar (good) performance when predicting the \emph{scene}.
The last row shows a failure case (note that CLIP also fails to give good noun captions in this case).
It is interesting to see that \modelname{} correctly identifies the \emph{reacher} as a boy but assigns the wrong attribute to it.

\section{Limitations}
\label{sec:limitations}
\modelname{} has some limitations to address in future work:

\begin{enumerate}
    \item As seen in \cref{table:vc}, the VC may work well only on tasks similar to the PPT dataset.
However, the VC alone cannot solve the PPT task and the CLIP backbone is indeed improved, evident through our downstream experiments.
    \item The VidSitu dataset comprises movie clips, that are person-centric.
Thus, we expect \modelname{} to work well broadly on videos with people.
Please note that VidSitu is used as an example in our work, and our take-away message is that \emph{dense text prompts can efficiently improve representations}.
    \item Our paper proposes that PPT on a small, densely annotated dataset is effective.
However, we acknowledge that curating such a dataset is challenging and costly.
While recent VLM advances may offset some of these annotations, the upfront annotation costs offset the massive computation costs later.
\end{enumerate}

\end{document}